\documentclass{article}
\usepackage[dblblindworkshop, final]{neurips_2026}
\workshoptitle{The 6th Workshop on Mathematical Reasoning and AI}

\usepackage[utf8]{inputenc}
\usepackage[T1]{fontenc}
\usepackage{hyperref}
\usepackage{url}
\usepackage{booktabs}
\usepackage{amsmath}
\usepackage{amsfonts}
\usepackage{microtype}
\usepackage{xcolor}
\usepackage{enumitem}
\usepackage{xspace}

\hypersetup{
  hidelinks,
  pdftitle={ClaimScope: Label-Blind Witness Verification for Mathematical Claim Auditing},
  pdfauthor={Mushir Akhtar},
  pdfsubject={MATH-AI 2026 workshop paper},
  pdfkeywords={mathematical reasoning, claim auditing, counterexamples, verification},
  pdfcreator={},
  pdfproducer={}
}
\newcommand{\claimscope}{\textsc{ClaimScope}\xspace}
\definecolor{csGreen}{HTML}{A5CF83}
\definecolor{csYellow}{HTML}{F0E76F}
\definecolor{csGold}{HTML}{ECB65F}
\definecolor{csOrange}{HTML}{E89951}
\title{\claimscope: Label-Blind Witness Verification for Mathematical Claim Auditing}
\author{Mushir Akhtar\\\texttt{mushirakhtar.ml@gmail.com}}

\begin{document}
\maketitle

\begin{abstract}
Mathematical agents are commonly evaluated by answer accuracy or proof acceptance. We introduce \claimscope, a 72-task benchmark for auditing proposed claims through a verdict, scope, counterexample, and repair. Tasks cover optimizer-state transport, finite-horizon affine recurrences, and exact $2\times2$ linear algebra. We evaluate Qwen3-8B with direct generation and four matched two-pass conditions, each starting from the same draft. A witness checker evaluates only the model's submitted state, without consulting reference verdicts or supplying a replacement witness. Its feedback still
conveys information about validity: a verified refutation establishes that the claim is false, and rejection reasons can suggest structural fixes. In this evaluation, verifier feedback yields $23/39$ accepted counterexamples on invalid claims and $25/72$ safe completions, compared with $15/39$ and
$20/72$ for checker-free generic feedback. Strict completion remains $17/72$; verdict accuracy changes from $47/72$ to $46/72$. The claim-grouped safe-completion difference is $.094$, with a bootstrap
95\% interval of $[.000,.198]$. All seven task-level gains follow horizon/input-length diagnostics. A post-hoc schema-only correction of the direct outputs reaches the same 25 safe tasks without another model call. These results separate claim classification from executable evidence, while locating the observed gains in witness-structure correction rather than demonstrating broader reasoning improvement.
\end{abstract}

\section{Introduction}

Reliable mathematical agents need more than correct final answers. A derivation
may agree with several examples yet fail as a universal statement: it may use
the wrong quantifier, omit a singular case, or confuse equality of an update
direction with equality of a complete transition. An audit must therefore
determine not only whether a claim is valid, but also its justified scope, the
reason for any failure, and an appropriate repair.

For example, transferring state between optimizers can preserve the
gradient direction yet change the parameter update because the learning rates
or weight-decay factors differ. A useful audit identifies the missing condition,
exhibits a parameter state where the proposed identity fails, and specifies a
correction. Such checks can support review of derivations and implementations;
the small claims here model this local task, not an entire training pipeline.

Existing benchmarks emphasize complementary endpoints. MATH
\citep{hendrycks2021math} and GSM8K \citep{cobbe2021gsm8k} score problem
solutions, while miniF2F \citep{zheng2021minif2f}, LeanDojo
\citep{yang2023leandojo}, ProofNet \citep{azerbayev2023proofnet}, and
AlphaGeometry \citep{trinh2024alphageometry} target formalization or proof.
COUNTERMATH asks for proof or disproof through counterexamples
\citep{li2025countermath}, and MathDebugger diagnoses errors in synthetic
questions and solutions \citep{liang2025mathdebugger}.

In this work, we introduce \claimscope for structured audits of already-stated
local claims. Most closely, CRAFT searches numerically for counterexamples to
machine-learning and optimization claims and presents a found witness to
the model for revision \citep{liu2026craft}. Our checker instead validates
the witness proposed by the model. It neither searches for nor supplies a
replacement witness or repair. This distinction concerns who constructs the
evidence, not whether the feedback conveys information about the answer.
Recent systems combine language models with proof-assistant feedback or
self-correction \citep{ren2025deepseekproverv2,lin2025goedelproverv2,zhao2025spoc};
our evaluation unit is a local claim and a submitted refutation rather than
a complete proof.

We use \emph{label-blind} operationally: the checker does not consult the stored
reference verdict, repair, or diagnostic label when producing feedback.
It is \emph{not verdict-independent}. A verified counterexample logically
implies an invalid claim, whereas a rejected witness establishes neither truth
nor falsity. A diagnostic rejection can also reveal how to fix the witness.
This setting is related to counterexample-guided synthesis
\citep{jha2014cegis}, but checks the model's proposed state instead of
returning a new counterexample. We ask whether this feedback improves audits
relative to matched revision without a checker.

We contribute (i) a 72-task benchmark with computational counterexample
validation; (ii) a protocol for model-proposed witness checking with an explicit
information boundary; and (iii) a paired analysis of classification, executable
evidence, and complete structured audits. The evaluation includes constant
policies, claim-grouped uncertainty, and a schema-only analysis to distinguish
structural correction from stronger mathematical reasoning.
The benchmark, checking and scoring code, and recorded outputs are available
in the project repository.\footnote{\url{https://github.com/Anonymous-X123/Claim_Scope}}

\section{Benchmark and protocol}

\paragraph{Tasks and reference validation.}
The benchmark contains 24 tasks per family: optimizer-state transport,
affine recurrences, and linear algebra (Table~\ref{tab:tasks}).
There are 33 valid and 39 invalid claims.
Optimizer tasks vary learning-rate scaling, bias correction, stabilizer placement,
decay semantics, and the distinction between direction and complete update.
Recurrence tasks test finite-horizon unrolling, indexing, and the singular case
$\rho=1$. Matrix tasks test determinant, inverse, trace, and transpose identities.

Reference validity rests on algebraic identities under the stated assumptions,
not agreement on a finite diagnostic grid. For optimizer claims, symbolic
substitution of the candidate transport into the update gives either an
identically zero discrepancy or a refutable expression. Standard finite-sum
and matrix identities establish the other valid cases. Every invalid claim
has an admissible, computationally verified counterexample in the requested
schema. Appendix~\ref{app:reference} gives the validation rules and explicit
optimizer counterexamples. Numerical optimizer checks use relative tolerance
$10^{-10}$ and absolute tolerance $10^{-12}$; recurrence and matrix checks use
exact rational arithmetic.

\begin{table}[ht]
\centering
\caption{Benchmark composition. Claim groups collect prompts with the same
formal specification, including assumptions and evaluation target.}
\label{tab:tasks}
\small
\begin{tabular}{@{}lrrrl@{}}
\toprule
Family & Tasks & Valid / invalid & Groups & Witness check \\
\midrule
Optimizer transport & 24 & 9 / 15 & 24 & numerical updates \\
Affine recurrence & 24 & 12 / 12 & 16 & exact unrolling \\
Linear algebra & 24 & 12 / 12 & 8 & exact matrix identities \\
\midrule
Total & 72 & 33 / 39 & 48 & \\
\bottomrule
\end{tabular}
\end{table}

\paragraph{Response and scope.}
Each prompt supplies the claim, assumptions, allowed labels, and witness schema,
but no answer key. The response contains a task identifier, verdict, scope,
failure mode, counterexample, and repair. Scope denotes the designated
evaluation target---for example, an optimizer direction versus a complete
transition---not a free-form proof of the strongest possible theorem.
Repairs are identifiers for corrections defined in the prompt.
Table~\ref{tab:scoring} details their scoring. The following task illustrates
the full audit.

\begin{center}
\setlength{\fboxsep}{5pt}
\fcolorbox{csOrange}{csYellow!18}{\parbox{0.94\linewidth}{\small
\textbf{Claim and assumptions.} For $x_{k+1}=\rho x_k+u_k$ with constant
inputs, the finite-horizon value is
$\widetilde{x}_T=\rho^T x_0+\sum_{k=0}^{T-1}\rho^{T-k}u_k$.\\
\textbf{Audit.} Invalid; scope: finite-horizon value; failure: input exponent.\\
\textbf{Counterexample.} $\rho=2$, $x_0=1$, $T=2$, inputs $[1,1]$.
Iteration gives $x_1=3$, $x_2=7$; the claim gives $4+4+2=10$.\\
\textbf{Repair.} Use
$x_T=\rho^Tx_0+\sum_{k=0}^{T-1}\rho^{T-1-k}u_k$.}}
\end{center}

\paragraph{Scoring.}
Safe completion is a conjunction of structured-output validity, correct
identifier and verdict, reference scope and repair, and the witness condition:
\[
\mathrm{Safe}_i(y)=
\mathrm{JSON}_i(y)\wedge\mathrm{ID}_i(y)\wedge
\mathrm{Verdict}_i(y)\wedge\mathrm{Scope}_i(y)\wedge
\mathrm{Repair}_i(y)\wedge W_i(y).
\]
Here $W_i$ requires a verified refutation for an invalid claim and a null witness
for a valid claim. Strict completion additionally requires the reference
failure-mode label. Scope, repair, and failure-mode scores compare normalized
labels, not semantic equivalence of arbitrary text. Thus safe completion
means compliance with this audit specification; it is not a proof of every
field's unrestricted mathematical meaning. To avoid rewarding an always-null
witness policy, we separately report accepted counterexamples among invalid
claims and inappropriate non-null witnesses among valid claims.

\begin{table}[ht]
\centering
\caption{Interpretation of the response fields and their checks.}
\label{tab:scoring}
\small
\begin{tabular}{@{}p{.17\linewidth}p{.75\linewidth}@{}}
\toprule
Field & Check and limitation \\
\midrule
Verdict & Valid or invalid, compared with the algebraically audited reference. \\
Scope & Canonical target: optimizer direction or full transition, finite-horizon
value, or matrix identity. No open-ended theorem-strength comparison. \\
Repair & Canonical correction named in the prompt, or none for a valid claim.
Equivalent free-form formulas and alternative restrictions receive no automatic
equivalence credit. \\
Failure mode & Canonical diagnostic label. Required for strict, but not safe,
completion because different descriptions can identify the same defect. \\
Counterexample & For an invalid claim, parse the state, enforce assumptions,
and recompute both expressions. Reported expected/actual values are not
trusted. For a valid claim, require null. \\
\bottomrule
\end{tabular}
\end{table}

\paragraph{Revision and feedback.}
Direct generation makes one call. Each revision condition makes the same draft
call and one review call. Self-reflection requests a critical review;
counterexample-guided revision asks the model to seek a refutation;
generic feedback warns that the witness may be wrong while explicitly stating
that no checker supplies a task-specific signal. Verifier feedback instead
reports the result of checking the submitted state. These conditions compare
tool-assisted revision with checker-free self-feedback
\citep{madaan2023selfrefine}, rather than trained outcome or process supervision
\citep{cobbe2021gsm8k,lightman2023verify}.

The checker returns \emph{verified} for an admissible refutation,
\emph{rejected} for a malformed, inadmissible, or non-refuting witness, and
\emph{not applicable} when the draft does not request a witness check.
A rejection can name a validation error, including a mismatch between the
horizon and number of inputs. The model receives the claim, its draft, the
status and reason, and a fixed revision instruction. It receives no reference
record, replacement witness, repair, or proof. Although the verifier loads
hash-checked formal task sources separately from the model context, its
feedback function uses no reference labels. Reproducing all 72 saved feedback
messages with label-free records confirms this separation.

\section{Experiment}

\paragraph{Configuration and comparison.}
We evaluate Qwen3-8B \citep{qwen2025qwen3} with 4-bit NormalFloat quantization
on one NVIDIA RTX A4500, temperature zero, thinking disabled, and at most
512 new tokens per call. All five conditions use the same 72 prompts.
Every two-pass draft is byte-identical to its direct counterpart. Each revision
condition therefore uses 144 calls with the same token ceilings, against
72 calls for direct generation. The generic-feedback condition is the closest
checker-free control in budget and review wording. Appendix~\ref{app:config}
records the model revision, runtime, and provenance. All analyses below use
these saved outputs; the additional baselines require no model generation.

\paragraph{Claim-grouped result.}
We give equal weight to each of the 48 formal claim groups before averaging
safe completion, because multiple prompts can express the same claim.
The grouped score is $.205$ for generic feedback and $.299$ for verifier
feedback, a difference of $.094$. A 10,000-sample paired bootstrap over groups
gives a percentile 95\% interval of $[.000,.198]$. Six groups improve, one
declines, and 41 do not change. The interval includes zero, and groups can
still share formula components: these point estimates do not establish a
general improvement. Task-level counts in Table~\ref{tab:results} provide the
underlying descriptive results.

\begin{table}[!htbp]
\centering
\caption{Counts on the frozen task set. CE is counterexample-guided
revision. Cex counts verified witnesses on 39 invalid claims (higher is better);
Non-null counts inappropriate non-null witnesses on 33 valid claims (lower is
better). The last three rows are deterministic, model-free policies or
post-processing; they are not additional generation conditions.}
\label{tab:results}
\small
\setlength{\tabcolsep}{4pt}
\begin{tabular}{@{}lrrrrrr@{}}
\toprule
Condition & JSON & Verdict & Cex & Non-null & Safe & Strict \\
 & /72 & /72 & /39 & /33 & /72 & /72 \\
\midrule
Direct & 72 & 46 & 15 & 26 & 18 & 15 \\
Self-reflection & 70 & 44 & 12 & 26 & 15 & 13 \\
CE-guided & 72 & 46 & 14 & 26 & 17 & 14 \\
Generic feedback & 71 & 47 & 15 & 24 & 20 & 17 \\
Verifier feedback & 71 & 46 & 23 & 25 & 25 & 17 \\
\midrule
Always valid, null witness & 72 & 33 & 0 & 0 & 33 & 33 \\
Always invalid, null witness & 72 & 39 & 0 & 0 & 0 & 0 \\
Direct + schema-only rule & 72 & 46 & 23 & 26 & 25 & 17 \\
\bottomrule
\end{tabular}
\end{table}

\paragraph{Evidence and classification.}
Generic-to-verifier safe completion changes from $20/72$ to $25/72$, with
seven gains and two losses; strict completion remains $17/72$.
Accepted counterexamples on false claims increase from $15/39$ to $23/39$,
but verdict accuracy decreases from $47/72$ to $46/72$.
The direct draft calls all 39 invalid claims invalid, but also rejects 26 of
33 valid claims. Its classification bias therefore leaves substantial room
for improving true-claim recognition rather than merely finding more
refutations. Among invalid claims, safe completion increases from $11/39$
under generic feedback to $18/39$ with verification; among valid claims,
it decreases from $9/33$ to $7/33$.

The constant policies expose an important endpoint limitation.
An always-null witness satisfies the isolated witness condition on all
33 valid claims without constructing any evidence. An always-invalid verdict
scores $39/72$. More strongly, an always-valid policy that copies the
prompt's scope and identifier, emits null, and selects no repair or failure
obtains $33/72$ safe completions, exceeding every model condition.
Valid claims do not require a proof in this benchmark.
Safe completion therefore must be read together with class-conditional
counterexample and classification scores, not as a standalone measure of
reliable mathematical reasoning.

\paragraph{Where revision helps.}
Family-level safe counts for direct/generic/verifier feedback are
$4/4/4$ for optimizer transport, $1/3/8$ for recurrences, and $13/13/13$
for linear algebra, each out of 24. All seven gains over generic feedback
follow a rejection identifying a horizon/input-length mismatch; none follows
only a non-refutation message. These gains span six formal claim groups and
four recurrence formulas. Of 27 verifier-condition revisions, 26 change only
the counterexample field; the remaining revision produces no parseable audit.
No parseable revision changes the verdict.

\paragraph{Schema-only analysis.}
To assess how much the witness representation explains, we apply one post-hoc
rule to direct outputs: if a recurrence witness reports a horizon different
from its input-list length, set the horizon to that length and leave every
other field unchanged. The task quantifies over admissible horizons, so this
selects a different witness rather than changing the claim.
The rule uses only the public schema, no reference label or formula evaluation,
and changes 21 records. It reaches the same 25 safe tasks as verifier feedback,
including the same seven gains over direct generation.
This is not a prospective baseline or a binary-feedback ablation.
It shows that the observed safe-task set can be recovered by deterministic
structural correction; the data do not require a stronger explanation in
terms of mathematical search.

\section{Discussion and limitations}

The study isolates model-proposed witness validation from checker-driven
counterexample search, but not validity feedback from diagnostic hints.
A verified status entails falsity, and a specific schema error can suggest a
correction. In the saved drafts, all 15 verified states belong to invalid
claims; the 21 horizon/input-length rejections split into 12 valid and nine
invalid claims, while 29 non-refuting states split into 14 valid and 15
invalid claims. Thus rejection is not a disguised reference verdict, but the
channel is informative and should not be described as verdict-blind.
A binary-only condition would be needed to isolate the value of status alone.

The benchmark is small, synthetic, and limited to one quantized model with
a short deterministic output budget. Repeated formulas, even across distinct
assumptions, limit independence. The results neither establish behavior on
unseen mathematical families nor demonstrate open-ended theorem proving or
optimizer performance in training. Numerical optimizer validation is
tolerance-based, unlike the exact recurrence and matrix checks.
Canonical scope and repair labels also restrict the endpoint: an alternative
mathematically sound repair can fail the label comparison.
Algebraic validation of reference claims, explicit class-conditional metrics,
and the trivial-policy results are therefore essential to interpreting this
testbed. Larger independent claim sets, semantic repair checking, and
proof obligations for valid claims remain important directions.

\section{Conclusion}
\claimscope evaluates structured audits of mathematical claims and
separates verdict classification from executable counterexamples.
In this fixed Qwen3-8B evaluation, diagnostic verifier feedback yields more
accepted counterexamples and safe completions than matched generic feedback,
without better verdict or strict-completion scores. Grouped uncertainty
includes no improvement, and a schema-only rule recovers the same safe-task
set. The clearest result is a witness-structure correction effect, not evidence
that verification alone improves general mathematical reasoning.

\clearpage
\bibliographystyle{plainnat}
\bibliography{references}

\clearpage
\appendix
\section{Supplementary evaluation details}

\subsection{Reference validation and admissible counterexamples}
\label{app:reference}

For an optimizer state, write
\[
Q=\sqrt{V/(1-\beta_2^t)}+\epsilon,\qquad
A=\frac{M}{(1-\beta_1^t)Q},\qquad
W_{\rm M}=(1-\alpha\lambda_{\rm M})W-\alpha D.
\]
The AdamW transition is
$W_{\rm A}=(1-\eta\lambda_{\rm A})W-\eta A$.
Its discrepancy from the reference equals
\[
W_{\rm A}-W_{\rm M}
=(\alpha\lambda_{\rm M}-\eta\lambda_{\rm A})W
+\alpha D-\eta A.
\]
Substituting
$M=(1-\beta_1^t)(\alpha/\eta)DQ$ gives
$A=(\alpha/\eta)D$, hence equal step directions, but leaves the first
decay term. Matching the full transition requires either matched decay
or the transport
\[
M=(1-\beta_1^t)
\left[\frac{\alpha}{\eta}D+
\left(\frac{\alpha\lambda_{\rm M}}{\eta}-\lambda_{\rm A}\right)W\right]Q.
\]
For a direction claim, the comparison is
$A=(\alpha/\eta)D$, equivalently $\eta A=\alpha D$.
For protocols that explicitly use the reference decay, that decay replaces
AdamW decay in the discrepancy.

The admissible optimizer domain enforces finite values, $V\geq0$,
positive learning rates, $0<\beta_1<1$, $0\leq\beta_2<1$,
integer $t\geq1$, $\epsilon\geq0$, and $Q>0$, together with the prompt's
extra assumptions.
The audit substitutes each candidate formula and its stated constraints into
the appropriate discrepancy. All nine valid optimizer specifications reduce
identically to zero; each of the other 15 has a verified admissible state.
This universal-identity check is distinct from testing a finite grid.

The following three cases show why accidental numerical agreement is
insufficient. Use $W=D=V=1$, $\beta_1=.9$, $\beta_2=.999$, $t=1$,
$\epsilon=0$, $\alpha=.1$, and $\eta=.01$.
\begin{itemize}[leftmargin=1.5em,itemsep=0.3em]
\item \texttt{PILOT-OPT-002}: with $\lambda_{\rm M}=.1$ and
$\lambda_{\rm A}=.01$, the direction transport gives $A=10$.
The complete updates are $W_{\rm A}=.8999$ and $W_{\rm M}=.89$.
The stipulated $\epsilon=0,V>0$ does not match decay. The reference verdict is
invalid; the canonical failure is decay mismatch and the repair is full transport.
\item \texttt{PILOT-OPT-013}: choose $\lambda_{\rm M}=.01$ and
$\lambda_{\rm A}=.1$, so $\alpha\lambda_{\rm M}=\eta\lambda_{\rm A}=.001$.
The candidate omits $\alpha/\eta$, giving $A=1$ rather than $10$.
Thus $\eta A=.01\ne.1=\alpha D$ despite matched decay.
The reference verdict is invalid; the failure is learning-rate scaling and
the repair is direction transport.
\item \texttt{PILOT-OPT-024}: with $\lambda_{\rm M}=.1$ and
$\lambda_{\rm A}=.01$, reversed decay compensation gives
$A=10+(.01-1)=9.01$ and $W_{\rm A}=.9098\ne.89=W_{\rm M}$.
The reference verdict is invalid; the failure is decay compensation and
the repair uses the full transport with the sign shown above.
\end{itemize}

For recurrences, induction gives
$x_T=\rho^T x_0+\sum_{k=0}^{T-1}\rho^{T-1-k}u_k$.
At $\rho=1$ this becomes $x_0+\sum_k u_k$; for constant input $u$
and $\rho\ne1$, the sum is $u(1-\rho^T)/(1-\rho)$.
The valid specifications use these identities only under their required
conditions or handle $\rho=1$ piecewise.
For matrices, the valid specifications are the reversed-order transpose,
cyclic trace, determinant product, and reversed-order inverse identities;
inverse checks require the relevant matrices to be invertible.
Invalid specifications in both families have exact rational witnesses.
Across all families, the 39 invalid claims admit witnesses in the published
schema; representability therefore does not impose a lower ceiling.

\subsection{Recorded worked example}
\label{app:worked-example}

The main-text example is \texttt{PILOT-REC-023}. Its shared draft labels
the input-exponent claim invalid and selects the exact-unroll repair, but
reports horizon $2$ with only one input, \texttt{[1]}.
The checker returns rejected with the local reason
\emph{horizon must equal the number of supplied inputs}.
This diagnostic does not specify which field to change, but it directly
identifies the inconsistent pair of fields. The model changes the input list
to \texttt{[1,1]}, retaining $\rho=2$, $x_0=1$, and horizon $2$.
The resulting $7\ne10$ comparison verifies the refutation.
The other audit fields do not change.
In the schema-only analysis, the alternative change to horizon $1$ also
refutes this claim: iteration gives $3$, while the candidate gives $4$.
Thus the model's final witness is not the only admissible structural correction.

\subsection{Scoring contract}
\label{app:scoring}

The scorer extracts the final parseable audit, compares its task identifier
with the originating task, and checks the five required audit fields.
Additional JSON keys are not rejected by this implementation.
A misspelled reported identifier receives zero identifier credit without
changing which benchmark task is scored. Scope, repair, verdict, and failure
labels are stripped of surrounding whitespace and lowercased before comparison.
The score does not trim spaces inside JSON key names or infer missing fields.

Consequently, reference agreement for scope and repair should not be read as
unrestricted semantic assessment. A verified witness does not certify the repair,
and the chosen canonical repair need not be the only mathematical correction.
Optimizer witnesses use scalar states, recurrences use rational initial state
and input lists, and matrix witnesses use rational $2\times2$ arrays.
Reported numerical values or diagnostic prose cannot substitute for the state.

The scorer's isolated witness flag treats a missing witness field like null
on valid claims. Safe completion still requires parseable structured output,
so this behavior cannot make an empty response safe.
We avoid interpreting that aggregate flag as evidence quality:
Table~\ref{tab:results} instead reports verified witnesses on invalid claims
and non-null witnesses on valid claims.
For verifier feedback, seven valid-claim responses explicitly contain null,
25 contain non-null witnesses, and one has no parseable audit.
For generic feedback these counts are nine, 24, and zero.
An always-null policy satisfies the isolated witness requirement on $33/72$
tasks, but verifies no invalid claim.

\begin{table}[ht]
\centering
\caption{Safe completion separated by reference verdict and family.}
\label{tab:classsafe}
\small
\begin{tabular}{@{}lrrrrr@{}}
\toprule
Condition & Invalid /39 & Valid /33 & Optimizer /24 & Recurrence /24 & Matrix /24 \\
\midrule
Direct & 11 & 7 & 4 & 1 & 13 \\
Self-reflection & 8 & 7 & 3 & 1 & 11 \\
CE-guided & 10 & 7 & 4 & 1 & 12 \\
Generic feedback & 11 & 9 & 4 & 3 & 13 \\
Verifier feedback & 18 & 7 & 4 & 8 & 13 \\
\bottomrule
\end{tabular}
\end{table}

\subsection{Feedback information and revision traces}
\label{app:feedback}

A verified status is logically sufficient for an invalid verdict.
An unparseable draft receives an unavailable status; this branch does not
occur in the 72 shared drafts. The checker never draws the explicit reference label, repair, failure mode,
or stored witness into the review message.
All saved feedback messages are reproducible after removing those reference
fields and retaining only the task specification.
This implementation-level separation does not imply statistical independence:
Table~\ref{tab:feedback} quantifies the actual relationship with validity.
In particular, the not-applicable cases already contain valid draft verdicts
with null witnesses; the checker does not itself certify those claims.

\begin{table}[ht]
\centering
\caption{Draft feedback stratified by audited reference verdict.
Non-refutation reports that the submitted state does not disprove the claim
under its assumptions.}
\label{tab:feedback}
\small
\begin{tabular}{@{}lrrr@{}}
\toprule
Feedback & Valid claim & Invalid claim & Total \\
\midrule
Verified refutation & 0 & 15 & 15 \\
Rejected: horizon/input-length mismatch & 12 & 9 & 21 \\
Rejected: non-refutation & 14 & 15 & 29 \\
Not applicable & 7 & 0 & 7 \\
\bottomrule
\end{tabular}
\end{table}

Among optimizer, recurrence, and matrix drafts respectively, the
verified/rejected/not-applicable counts are $8/16/0$, $1/23/0$, and $6/11/7$.
Final predictions change on $1/24$, $21/24$, and $5/24$ tasks.
The optimizer change is a parse failure; the other 26 changes affect only
the witness. Therefore no well-formed final revision changes a draft verdict.
All 39 truly invalid claims already have invalid draft verdicts.
Among these, accepted witnesses increase from 15 to 23.
Among the 26 valid claims incorrectly called invalid in the draft, no
verifier-condition response corrects the verdict.

The seven safe gains over generic feedback are
\texttt{PILOT-REC-002}, \texttt{005}, \texttt{007}, \texttt{009},
\texttt{010}, \texttt{012}, and \texttt{023}.
All follow horizon/input-length rejection and cover input-exponent,
initial-exponent, $\rho=1$ linear, and constant-input geometric formulas.
The two safe losses, \texttt{PILOT-REC-004} and \texttt{017}, are valid
piecewise-geometric claims that generic feedback accepts but verifier feedback
continues to reject. No gain over generic feedback follows only a
non-refutation message. This is descriptive attribution to message types,
not an experiment separating the effects of diagnosis and status.

\begin{table}[ht]
\centering
\caption{Verdict confusion: reference class $\to$ reported class.
V and I denote valid and invalid; M denotes missing or unparseable verdict.}
\label{tab:confusion}
\small
\begin{tabular}{@{}lrrrrrr@{}}
\toprule
Condition & V$\to$V & V$\to$I & V$\to$M & I$\to$V & I$\to$I & I$\to$M \\
\midrule
Direct & 7 & 26 & 0 & 0 & 39 & 0 \\
Self-reflection & 7 & 26 & 0 & 0 & 37 & 2 \\
CE-guided & 7 & 26 & 0 & 0 & 39 & 0 \\
Generic feedback & 9 & 24 & 0 & 1 & 38 & 0 \\
Verifier feedback & 7 & 25 & 1 & 0 & 39 & 0 \\
\bottomrule
\end{tabular}
\end{table}

\subsection{Paired and grouped results}

\begin{table}[ht]
\centering
\caption{Task transitions from generic feedback to verifier feedback.
The witness row includes only the 39 invalid claims; other rows include all 72.}
\label{tab:paired}
\small
\begin{tabular}{@{}lrrrr@{}}
\toprule
Metric & $0\to0$ & $0\to1$ & $1\to0$ & $1\to1$ \\
\midrule
Verified invalid-claim witness & 16 & 8 & 0 & 15 \\
Safe completion & 45 & 7 & 2 & 18 \\
Strict completion & 53 & 2 & 2 & 15 \\
Verdict correct & 24 & 1 & 2 & 45 \\
\bottomrule
\end{tabular}
\end{table}

A claim group is the serialized formal specification with task identifiers,
surface wording, and reference outputs excluded.
The 48 groups comprise 24 optimizer specifications, 16 recurrence
specifications, and eight matrix identities. Each matrix identity has three
wordings. Different groups can still share formulas and differ only in
assumptions or evaluation target, so grouping does not establish independence.
Within each group we average prompt scores; the reported overall grouped
score is the unweighted mean of these group averages.

\begin{table}[!htbp]
\centering
\caption{Equal-weight means over 48 formal claim groups.}
\label{tab:grouped}
\small
\begin{tabular}{@{}lrrr@{}}
\toprule
Condition & Verdict & Safe & Strict \\
\midrule
Direct & .632 & .184 & .122 \\
Self-reflection & .590 & .149 & .108 \\
CE-guided & .632 & .177 & .115 \\
Generic feedback & .642 & .205 & .142 \\
Verifier feedback & .632 & .299 & .153 \\
Always valid & .417 & .417 & .417 \\
Always invalid & .583 & .000 & .000 \\
Direct + schema-only rule & .632 & .299 & .153 \\
\bottomrule
\end{tabular}
\end{table}

For the paired safe difference, we sample 48 groups with replacement
10,000 times using seed 20260906, retaining generic and verifier outcomes
together. The percentile interval is $[.000,.198]$ around a mean difference
of $.09375$. Six groups improve, one declines, and 41 remain unchanged.
No independent-task significance claim follows from these data.
The schema-only rule is chosen after inspecting the saved feedback, not
prespecified. It modifies horizon to match input-list length without running a
checker or consulting a label. It produces the same set of 25 safe task
identifiers as verification, although the final witnesses need not be equal.

\subsection{Configuration, cost, and reproducibility}
\label{app:config}

\begin{table}[!htbp]
\centering
\caption{Inference configuration shared by the five generation conditions.}
\small
\begin{tabular}{@{}lp{.73\linewidth}@{}}
\toprule
Component & Setting \\
\midrule
Model & Qwen/Qwen3-8B \\
Revision & \texttt{b968826d9c46dd6066d109eabc6255188de91218} \\
Quantization & 4-bit NormalFloat; model thinking mode off \\
Decoding & Temperature 0; at most 512 new tokens per call \\
Hardware & One NVIDIA RTX A4500 \\
Runtime & PyTorch 2.12.1+cu126; Transformers 5.16.1; CUDA 12.6 \\
Draft control & Byte-identical direct and first-pass outputs for all 72 tasks \\
Integrity & Public task-file hash and reference-file hash checked separately \\
\bottomrule
\end{tabular}
\end{table}

\begin{table}[!htbp]
\centering
\caption{Recorded inference cost. Seconds are summed host-measured
generation wall times, not GPU-kernel time; model loading is excluded.}
\small
\begin{tabular}{@{}lrrr@{}}
\toprule
Condition & Model calls & Output tokens & Seconds \\
\midrule
Direct & 72 & 8,719 & 853.9 \\
Self-reflection & 144 & 17,457 & 1,757.5 \\
CE-guided & 144 & 17,418 & 1,744.9 \\
Generic feedback & 144 & 17,367 & 1,772.3 \\
Verifier feedback & 144 & 17,436 & 1,672.4 \\
\bottomrule
\end{tabular}
\end{table}

The task file is unchanged throughout these comparisons. Its SHA-256 hash is
\begin{center}\footnotesize
\texttt{87bd5117f13a4335a950d858343ca02dc7d9390a765d2495772e7ef827ed13ec}.
\end{center}
Hash verification establishes artifact identity, not mathematical correctness.
The reference-validation analysis explicitly supplies invalid labels and
witnesses for \texttt{PILOT-OPT-002}, \texttt{013}, and \texttt{024};
the public prompts and recorded responses remain fixed.
The separate audited reference contains 33 valid and 39 invalid claims.
Feedback reproduction with label-free records confirms that rescoring does not
alter the messages the model actually receives.

The accompanying \texttt{scripts/reanalyze.py} script reads
the saved five-condition outputs, checks shared drafts, audits reference
specifications, and writes per-task scores, class-conditional counts,
grouped comparisons, deterministic baselines, and a provenance file.
It takes an optional \texttt{--artifact-root} and performs no model inference.
It records the input, scorer, and analysis-script hashes. The original
reference is preserved alongside the audited reference so that the label
changes and every aggregate can be traced to individual tasks.
The computational checks support this finite benchmark; they do not constitute
a general proof system for arbitrary mathematical text.

\end{document}